\documentclass[10pt,journal,compsoc]{IEEEtran}
\usepackage{flushend}
\ifCLASSOPTIONcompsoc
  \usepackage[nocompress]{cite}
\else
  \usepackage{cite}
\fi

\ifCLASSINFOpdf
   \usepackage[pdftex]{graphicx}
\else
\fi

\ifCLASSINFOpdf
\else
\fi

\usepackage{multirow}
\usepackage{makecell}
\usepackage[cmintegrals]{newtxmath}
\usepackage{bm}
\usepackage{algorithm}
\usepackage{algorithmic}
\usepackage{indentfirst}
  
\usepackage{amsmath,amssymb}
\usepackage{array}
\usepackage{pifont}
\usepackage{mathrsfs}
\usepackage{cite}
\usepackage{tikz}
\usepackage{color}
\usepackage{amsmath}
\usepackage{colortbl}
\usepackage{bbding}

\newcommand*{\circled}[1]{\lower.7ex\hbox{\tikz\draw (0pt, 0pt)%
    circle (.5em) node {\makebox[1em][c]{\small #1}};}}

\ifCLASSOPTIONcompsoc
 \usepackage[caption=false,font=normalsize,labelfont=sf,textfont=sf]{subfig}
\else
 \usepackage[caption=false,font=footnotesize]{subfig}
\fi

\begin{document}
\title{MoE-FFD: Mixture of Experts for Generalized and Parameter-Efficient Face Forgery Detection}

\author{Chenqi~Kong,~\IEEEmembership{Member,~IEEE}, Anwei~Luo, Peijun Bao, Yi Yu, Haoliang~Li,~\IEEEmembership{Member,~IEEE}, Zengwei ~Zheng,
Shiqi~Wang,~\IEEEmembership{Senior Member,~IEEE}, and 
Alex C. Kot,~\IEEEmembership{Life Fellow,~IEEE}
\IEEEcompsocitemizethanks{\IEEEcompsocthanksitem C. Kong, P. Bao, Y. Yu, and A. C. Kot are with the Rapid-Rich Object Search (ROSE) Lab, School of Electrical and Electronic Engineering, Nanyang Technology University, Singapore, 639798. \protect\\
E-mail: (chenqi.kong@ntu.edu.sg, peijun001@e.ntu.edu.sg, yuyi0010@e.ntu.edu.sg, eackot@ntu.edu.sg)
\IEEEcompsocthanksitem A. Luo (Corresponding Author) is with the School of Computing and Artificial Intelligence, Jiangxi University of Finance and Economics, Nanchang 330013, China, and also with Jiangxi Provincial Key Laboratory of Multimedia Intelligent Processing, Nanchang 330032, China. \protect\\
E-mail: luoanw@mail2.sysu.edu.cn

\IEEEcompsocthanksitem H. Li is with the Department
of Electrical and Engineering, City University of Hong Kong, Hong Kong, China. \protect\\
E-mail: haoliang.li@cityu.edu.hk
\IEEEcompsocthanksitem Z. Zheng is with the Department of Computer Science and Computing, Zhejiang University City College, Zhejiang, China.  \protect\\
E-mail: zhengzw@zucc.edu.cn
\IEEEcompsocthanksitem S. Wang is with the Department
of Computer Science, City University of Hong Kong, Hong Kong, China. \protect\\
E-mail: shiqwang@cityu.edu.hk
}}

\markboth{Submitted to IEEE TRANSACTIONS ON DEPENDABLE AND SECURE COMPUTING}%
{Shell \MakeLowercase{\textit{et al.}}: Bare Demo of IEEEtran.cls for Computer Society Journals}

\IEEEtitleabstractindextext{%
\begin{abstract}
Deepfakes have recently raised significant trust issues and security concerns among the public. Compared to CNN-based face forgery detectors, ViT-based methods take advantage of the expressivity of transformers, achieving superior detection performance. However, these approaches still exhibit the following limitations: (1) Fully fine-tuning ViT-based models from ImageNet weights demands substantial computational and storage resources; (2) ViT-based methods struggle to capture local forgery clues, leading to model bias; (3) These methods limit their scope on only one or few face forgery features, resulting in limited generalizability. To tackle these challenges, this work introduces Mixture-of-Experts modules for Face Forgery Detection (MoE-FFD), a generalized yet parameter-efficient ViT-based approach. MoE-FFD only updates lightweight Low-Rank Adaptation (LoRA) and Adapter layers while keeping the ViT backbone frozen, thereby achieving parameter-efficient training. Moreover, MoE-FFD leverages the expressivity of transformers and local priors of CNNs to simultaneously extract global and local forgery clues. Additionally, novel MoE modules are designed to scale the model's capacity and smartly select optimal forgery experts, further enhancing forgery detection performance. Our proposed learning scheme can be seamlessly adapted to various transformer backbones in a plug-and-play manner. \textcolor{black}{Extensive experimental results demonstrate that the proposed method achieves state-of-the-art face forgery detection performance with significantly reduced parameter overhead in cross-dataset, cross-manipulation, and robustness evaluations. Our ablation studies further validate the effectiveness of the designed components and the proposed learning scheme.} The code is available at: https://github.com/LoveSiameseCat/MoE-FFD.

\end{abstract}

}

\maketitle
\IEEEdisplaynontitleabstractindextext
\IEEEpeerreviewmaketitle

\IEEEraisesectionheading{\section{Introduction}\label{sec:introduction}}

\IEEEPARstart{W}{ith} rapid advancements in Artificial Intelligence-Generated Content (AIGC), forged facial content has become increasingly sophisticated, making it difficult for the human eye to distinguish between fake and real faces. Non-experts can easily use face manipulation algorithms to create highly realistic falsified facial images and videos, known as Deepfakes. Consequently, the rapid proliferation of Deepfake content on social media platforms has led to significant security issues, including disinformation, fraud, and impersonation \cite{kong2022digital}. Even worse, the complexity of deployment environments in real-world applications further deteriorates the performance of detection models. Therefore, developing generalized and robust face forgery detectors to counter malicious attacks remains a substantial challenge.

\begin{figure}[ht]
\centering
\includegraphics[scale=0.36]{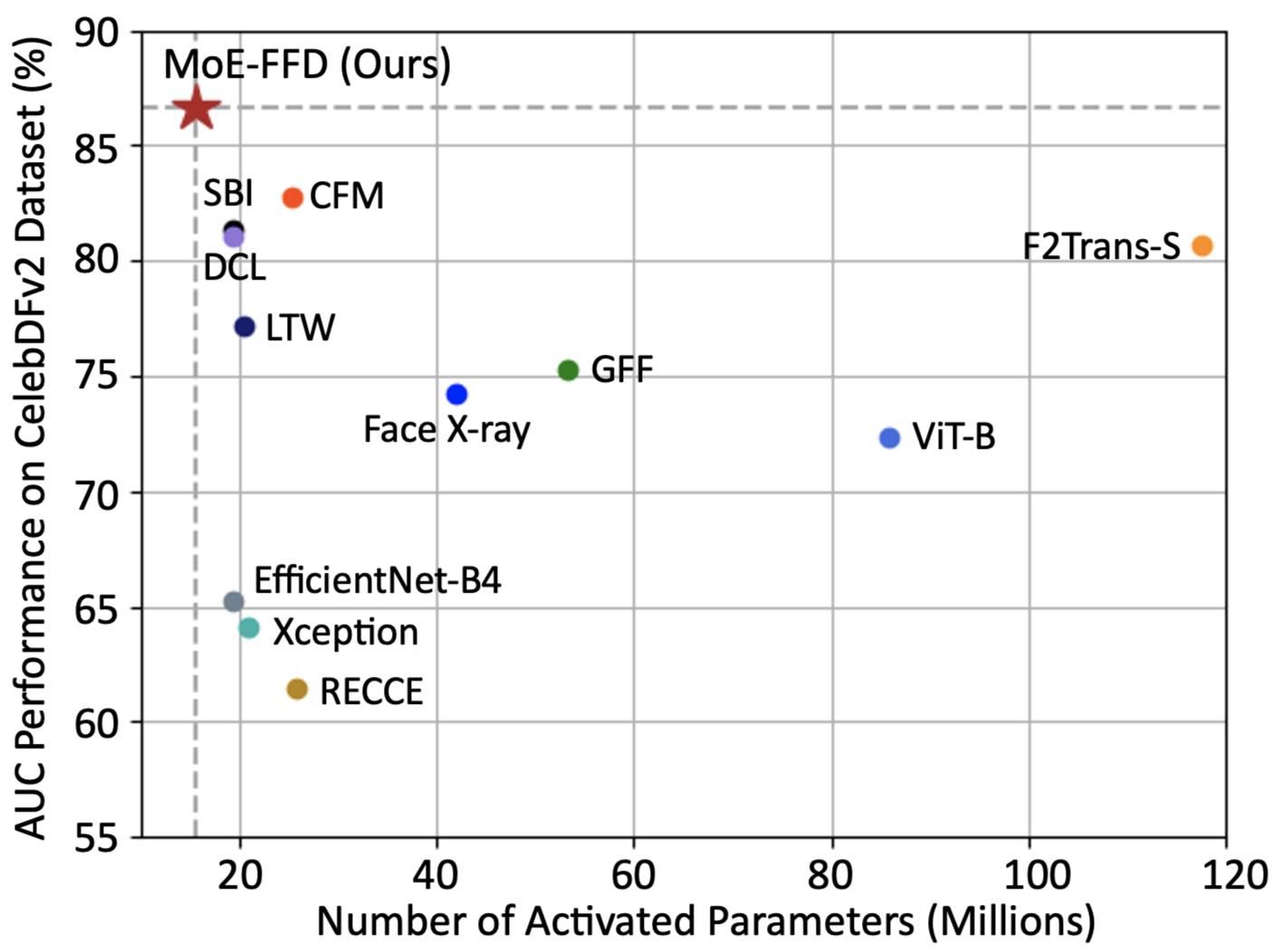}
\caption{Comparison between MoE-FFD (Ours) and Open-source face forgery detection models on the CelebDF-v2 dataset. We present the number of activated parameters and the AUC detection performance in $x$ and $y$ axis, respectively.}
\label{Teaser}
\end{figure}

Early traditional Deepfake detection methods focused on extracting hand-crafted features, such as eye-blinking frequency \cite{li2018ictu} and headpose inconsistency \cite{yang2019exposing}. These techniques fail to capture representative features because the hand-crafted features may limit their scopes to only one or few kinds of statistical information. Heading into the era of deep learning, numerous CNN-based methods have been proposed to improve the detection accuracy \cite{li2020face, dong2023implicit, 10315169, kong2021appearance, huang2023implicit}. Many of them employ Xception network \cite{chollet2017xception} or EfficientNet \cite{tan2019efficientnet} as backbones due to their outstanding performance in face forgery detection. To enhance generalizability and robustness, some methods proposed to extract common forgery features such as noise patterns \cite{luo2021generalizing, masi2020two, kong2022detect}, blending artifacts \cite{li2020face, shi2020informative}, frequency fingerprints  \cite{miao2022hierarchical, xu2020learning, li2021frequency, qian2020thinking}, and identity inconsistency  \cite{huang2023implicit}, etc. However, these methods are inherently limited to the local-interactions of CNN architectures.

With the advent of the vision transformer (ViT) \cite{dosovitskiy2020image}, ViT architectures have achieved significant success in a wide variety of computer vision tasks due to their long-range interactions and outstanding expressivity. In the realm of Deepfake detection, numerous ViT-based approaches \cite{dong2022protecting, shao2022detecting, zhuang2022uia, wang2022m2tr, kong2023enhancing, guan2022delving, miao2023f} have been proposed, achieving enhanced accuracy and generalizability. 
For instance, Dong et al. \cite{dong2022protecting} designed a ViT to capture the identity inconsistency between inner and outer face region. Zhuang et al. \cite{zhuang2022uia} proposed a ViT-based model to mine unsupervised inconsistency-aware features within each frame. Shao et al. \cite{shao2022detecting} introduced a concise yet effective Seq-DeepFake Transformer to predict a sequential vector of facial manipulation operations. 

Nevertheless, ViT-based forgery detection approaches still face several limitations. First, fully training ViT-based models from the ImageNet weights demands substantial computational resources, which hinders their deployment or fine-tuning in real-world applications, especially on resource-constrained mobile devices. Second, although ViT-based methods exhibit outstanding expressivity, they may struggle to capture forgery features in local abnormal regions, resulting in model bias and limited generalizability. Finally, previous methods focus on certain forgery artifacts, but it is challenging to empirically select the optimal features in unpredictable application scenarios.

This paper presents a generalized yet parameter-efficient approach MoE-FFD, which proposes using \textbf{M}ixture \textbf{o}f \textbf{E}xperts for \textbf{F}ace \textbf{F}orgery \textbf{D}etection. MoE-FFD draws inspiration from Parameter Efficient Fine-Tuning (PEFT), which integrates lightweight Low-Rank Adaptation (LoRA) layers and Adapter layers with the ViT backbone. During the training process, only the designed LoRA and Adapter parameters are updated while the ViT parameters remain frozen at their initial ImageNet weights. The designed MoE modules dynamically select optimal LoRA and Adapter experts for face forgery detection.

Compared with previous arts that directly fine-tune the entire ViT parameters, the proposed fine-tuning strategy effectively preserves abundant knowledge from ImageNet and enables the model to adaptively learn forgery-specific features. Additionally, the designed LoRA layers model the long-range interactions within input faces, while the Convpass Adapter layers effectively highlight local forgery anomalies. To this end, the integration of the designed LoRA and Adapter layers leverages the expressivity of transformers and the local forgery priors of CNNs, leading to enhanced generalizability and robustness. Additionally, we design novel Mixture of Experts (MoE) modules within both LoRA and Adapter layers, enabling the model to scale its capacity efficiently by activating only a subset of parameters dynamically. The MoE modules adaptively select optimal forgery detection experts based on input faces, further improving detection performance.

As depicted in Fig.~\ref{Teaser}, our MoE-FFD with the fewest activated parameters achieves the best AUC score on the unseen CelebDF-v2 dataset \cite{li2020celeb}. Overall, the contributions of this work are summarized as follows:
\begin{itemize}
    \item We innovatively integrate LoRA and Adapter modules with the ViT backbone for face forgery detection. This design enables the model to simultaneously mine global and local forgery clues in a parameter-efficient manner. 
    \item We design novel Mixture-of-Experts (MoE) modules to scale up the model capacity. These modules dynamically select optimal forgery experts for input faces, thereby boosting detection performance. Furthermore, the MoE modules can be seamlessly adapted to other transformer architectures in a plug-and-play fashion. 
    \item We conduct experiments on seven Deepfake datasets and various common perturbations. Our experimental results demonstrate that MoE-FFD achieves state-of-the-art generalizability and robustness. Extensive ablation experiments validate the effectiveness of our proposed MoE learning scheme and the designed components. 
\end{itemize}

The remainder of this paper is structured as follows. Section 2 comprehensively reviews previous related literature. Section 3 details the proposed mixture of expert framework for face forgery detection. Section 4 presents extensive experimental results under various settings. Finally, Section 5 concludes the paper and discusses potential directions for future research.

\begin{figure*}[ht]
\centering
\includegraphics[scale=0.37]{ 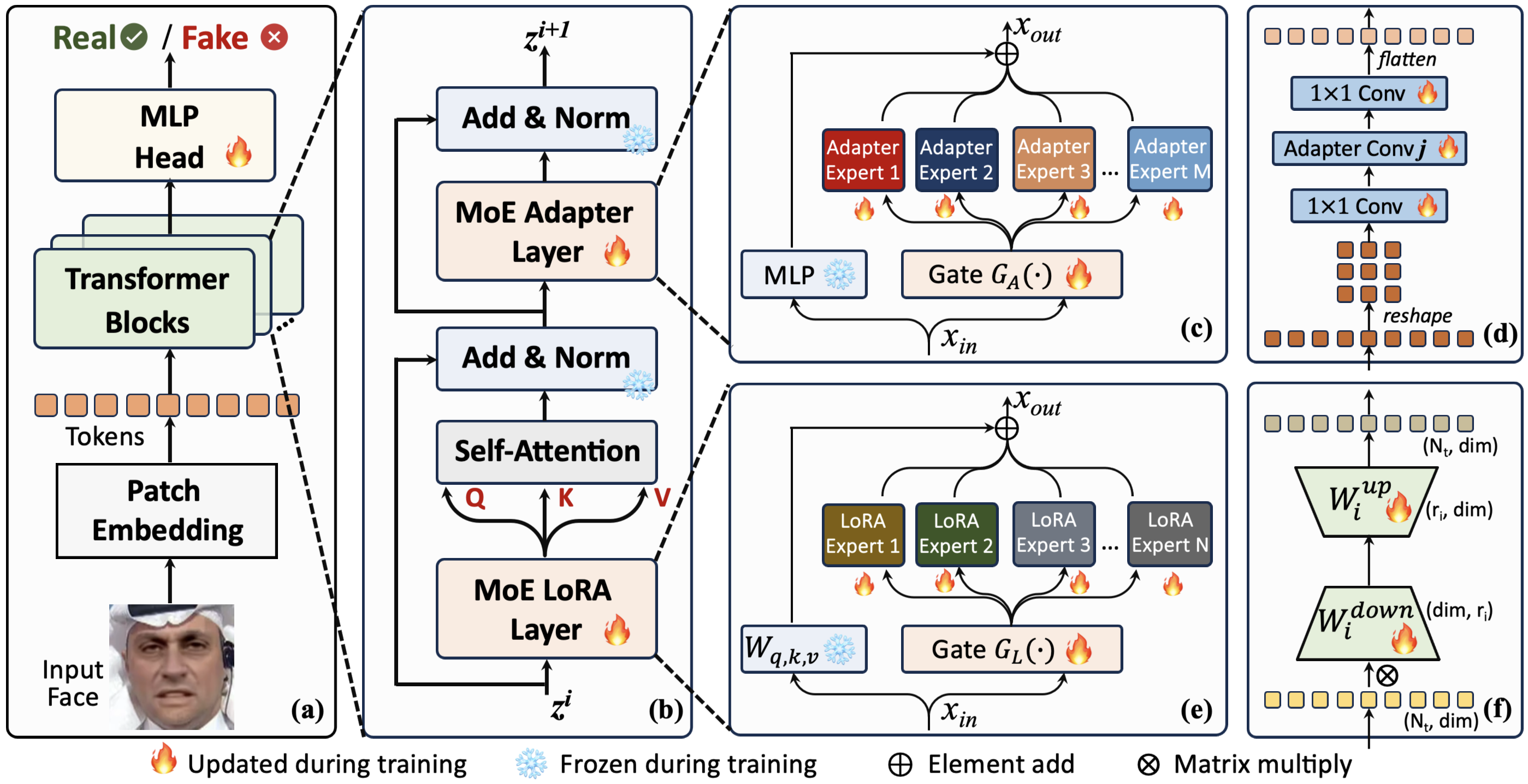}
\caption{Overview of the designed MoE-FFD framework. (a) Overall model structure; (b) Details of MoE-FFD transformer block; (c) Details of the designed MoE Adapter layer; (d) Details of each Adapter expert; (e) Details of the designed MoE LoRA layer; (f) Details of each LoRA expert. 
}
\label{Framework}
\end{figure*}

\section{Related Works}
In this section, we broadly review existing works
on face forgery detection, parameter-efficient fine-tuning, and mixture of experts. 

\subsection{Face Forgery Detection}
Early attempts at face forgery detection primarily relied on extracting handcrafted features, such as the lack of eye blinking \cite{li2018ictu}, inconsistency of head pose \cite{yang2019exposing}, face warping artifacts \cite{li2018exposing}, and heart rate anomalies \cite{ciftci2020fakecatcher}. However, these methods suffer from limited accuracy due to their narrow focus on specific statistical information. In response, learning-based approaches have emerged, leveraging generic network architectures like Xception Network \cite{chollet2017xception}, EfficientNet \cite{tan2019efficientnet}, and Capsule Network \cite{nguyen2019capsule} for forgery feature extraction. Nonetheless, CNN-based methods are prone to overfitting to the training data, resulting in limited generalizability and robustness \cite{deng2024towards, yu2024benchmarking}. Follow-up works, such as Face X-ray \cite{li2020face}, F$^{3}$Net \cite{qian2020thinking}, SBI \cite{shiohara2022detecting}, DCL \cite{sun2022dual}, RECCE \cite{cao2022end}, and MVIM \cite{zhang2024deepfake}, have introduced comprehensive forgery frameworks and robust feature extraction techniques, enhancing the model's generalization capability. ADA-FInfer \cite{hu2024ada} and Diff-ID \cite{yu2024diff} take one step forward to achieve explainable Deepfake detection. \textcolor{black}{Recent LVLM-based works \cite{huang2024ffaa, jia2024can, yu2025unlocking, zhang2025mfclip, guo2025rethinking} aim to achieve explainable face forgery detection.}
With the explosive development of ViT \cite{dosovitskiy2020image}, numerous ViT-based methods \cite{dong2022protecting, shao2022detecting, zhuang2022uia, wang2022m2tr, kong2023enhancing, guan2022delving, miao2023f, zhang2025mfclip} have been proposed to tackle the Deepfake problem. These approaches take advantage of non-inductive bias and global context understanding, achieving superior face forgery detection performance. However, most of these methods fine-tune the entire ViT model on Deepfake datasets, which is computationally expensive and may lead to loss of valuable knowledge from the ImageNet dataset \cite{deng2009imagenet}.
In contrast, our approach finetunes external lightweight LoRA and Adapter parameters during training, keeping the ViT backbone fixed with ImageNet weights, thus enabling
the model to learn forgery-specific knowledge and enhance the detection performance. 

\subsection{Parameter Efficient Fine-Tuning (PEFT) }
Over the past few years, deep learning models have increased exponentially in size, especially after the advent of transformers. Consequently, numerous Parameter Efficient Fine-Tuning (PEFT) methods have been proposed to reduce the computational and storage overhead. PEFT only updates a small subset of the model parameters while freezing the majority of pretrained weights. Adapter \cite{houlsby2019parameter, sung2022lst} is a typical PEFT method, comprising a down-sample layer and an up-sample layer, usually integrated into transformer layers and blocks. Low-Rank Adaptation (LoRA) \cite{hu2021lora} layer aims at updating two low-rank matrices, significantly reducing the trainable parameters. Zhong et al. \cite{zhong2024convolution} found that convolution operation effectively introduces local prior beneficial for image segmentation.  
Additionally, Visual Prompt Tuning (VPT) \cite{jia2022visual} augments inputs with extra learnable tokens, which can be regarded as learnable pixels to vision transformers. 
Neural Prompt Search (NOAH) \cite{zhang2022neural} advances further by incorporating Adapter, LoRA, and VPT into vision transformers and optimizing their design through a neural architecture search algorithm. Scale and Shift Feature Modulation (SSF) \cite{lian2022scaling} introduces to scale and shift parameters to modulate visual features during training, achieving comparable performance compared with full fine-tuning. Convpass \cite{jie2022convolutional} integrates convolutional bypasses into large ViT models, leveraging local priors to improve image classification performance. However, the application of PEFT methods to face forgery detection remains largely unexplored. 
In this work, we incorporate dedicated forgery Adapter and LoRA layers to mine global and local clues, thereby achieving a parameter-efficient face forgery detection.

\subsection{Mixture of Experts (MoE)}
Mixture-of-Experts (MoE) \cite{jacobs1991adaptive} models aim to augment the model's capacity without increasing computational expenses. MoE comprises multiple sub-experts and incorporates a gating mechanism to dynamically select the most relevant Top-k experts, thereby optimizing the results. The concept of MoE has been widely used in both Computer Vision \cite{riquelme2021scaling, lou2021cross, mustafa2022multimodal, shen2023scaling} and Natural Language Processing \cite{shazeer2017outrageously, lepikhin2020gshard, artetxe2021efficient}. Sparse MoE \cite{shazeer2017outrageously} introduces a router to select a subset of experts, ensuring that the inference time is on par with the standalone counterpart. Subsequent methods such as \cite{hazimeh2021dselect, lewis2021base, roller2021hash} seek to design novel gating mechanisms to enhance the performance on specific tasks. Follow-up works \cite{hazimeh2021dselect, kudugunta2021beyond, ma2018modeling} propose leveraging multi-task learning to guide the model to select optimal experts for a given input query. Moreover, some studies apply MoE architectures to domain adaptation \cite{guo2018multi} and domain generalization \cite{li2022sparse} tasks. Previous works usually adopt Feed-Forward Networks (FFN) as the expert choices \cite{shazeer2017outrageously, riquelme2021scaling, bao2022vlmo, du2022glam, zhou2022mixture, fedus2022switch}. However, Feed-Forward Networks (FFNs) still consume significant memory and computational resources during training and inference. In our work, we draw inspiration from the concept of Mixture-of-Experts (MoE) to automate the selection of different LoRA and Adapter modules for various test data. The use of MoE introduces negligible computational overhead, facilitating efficient Deepfake detection. Additionally, the designed MoE modules intelligently select optimal experts, significantly outperforming previous detection methods.

\section{Methodology}
Herein, we first briefly review the basic preliminaries on Vision Transformer (ViT), Low Rank Adaptation (LoRA), and Compass Adapter. Then we introduce the overview of our designed face forgery detection framework. Finally, we delve into the details of the designed architectures and objective functions of our method. 

\subsection{Preliminaries}
\subsubsection{Vision Transformer} Vision Transformer consists of multiple blocks of self-attention and Multi-Layer Perceptron (MLP). For a given input sequence $x \in \mathbb{R}^{N_{t} \times D}$, $x$ is firstly projected to queries $Q \in \mathbb{R}^{N_{t} \times dim}$, keys $K \in \mathbb{R}^{N_{t} \times dim}$, and values $V \in \mathbb{R}^{N_{t} \times dim}$ using three learnable matrices $W_{q} \in \mathbb{R}^{D \times dim}$, $W_{k} \in \mathbb{R}^{D \times dim}$, and $W_{v} \in \mathbb{R}^{D \times dim}$, where $N_{t}$, $D$, and $dim$ denote the token number, embedding dimension, and hidden dimension, respectively. The $Q$, $K$, and $V$ are calculated by:  
\begin{equation}
     Q = xW_{q}, K = xW_{k}, V = xW_{v}.
\end{equation}
Then the Self-Attention is conducted by:
\begin{equation}
     {\rm Attention}(Q,K,V)={\rm softmax}({QK^\top}/{\sqrt{dim}})V.
\end{equation}
\subsubsection{Low Rank Adaptation (LoRA)} LoRA is one popular parameter-efficient tuning method. For a pretrained ImageNet weight matrix $W \in \mathbb{R}^{D\times dim}$, LoRA freezes $W$ during training while adding a product of two trainable low-rank matrices $W^{down}W^{up}$, where $W^{down} \in \mathbb{R}^{D \times r}$, $W^{up} \in \mathbb{R}^{r \times dim}$, and rank $r << min(D, dim)$. As such, the forward pass is modified as: 
\begin{equation}
     h = x(W + \Delta W) = xW + xW^{down}W^{up},
\end{equation}
where $W$ could be either Query, Key, or Value matrix, i.e., $W \in {\{W_{q}, W_{k}, W_{v}\}}$. And $W$ is frozen during training. Compared with previous ViT-based finetuning methods, the proposed LoRA learning scheme significantly reduce the trainable parameters.

\subsubsection{Convpass Adapter} Adapter integrates local priors for visual tasks, constructing Convolutional Bypasses (Convpass) within the Vision Transformer (ViT) framework as adaptation modules. The Convpass Adapter can be formulated as:
\begin{equation}
     x_{out} = MLP(x_{in})+ Convpass(x_{in}),
\end{equation}
where the parameters in MLP are fixed with ImageNet weights during training. Convpass generally consists of several trainable convolutional layers. Compared with the MLP layer, the Convpass layer saves computational resources.

\begin{figure}[ht]
\centering
\includegraphics[scale=0.37]{ 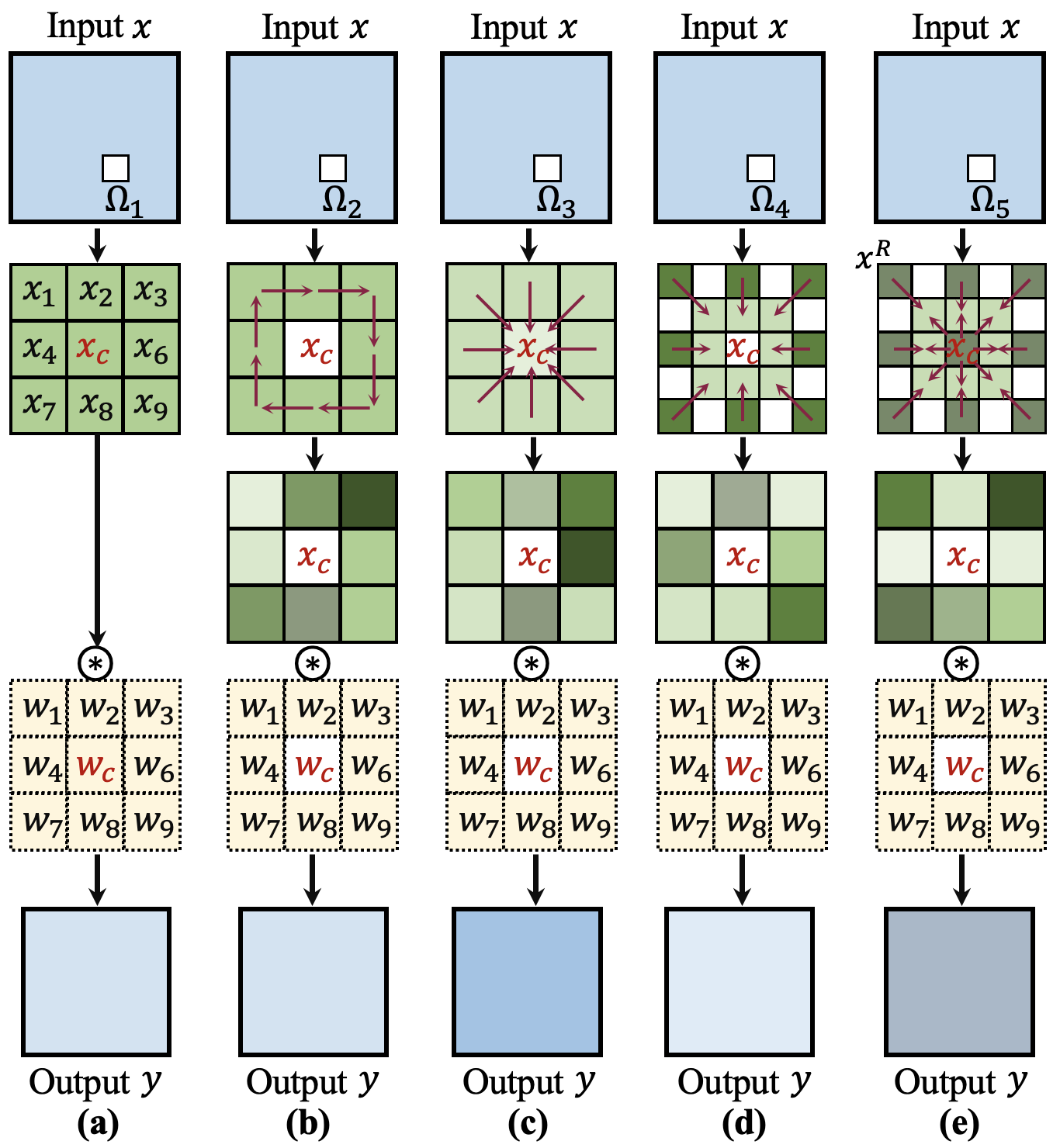}
\caption{Illustration of the designed Convpass Adapter experts. (a) Vanilla Convolution, (b) Angular Difference Convolution (ADC), (c) Central Difference Convolution (CDC), (d) Radial Difference Convolution (RDC), and (e) Second-Order Convolution
(SOC).
}
\label{MoE_adapter}
\end{figure}

\subsection{Overview of the Proposed Framework}
Fig.~\ref{Framework} illustrates the designed MoE-FFD framework. Fig.~\ref{Framework} (a) depicts the ViT backbone, which is initialized with the ImageNet weights. The input face is first processed through patch and positional embedding, after which the embedded tokens are passed into the customized transformer blocks. Fig.~\ref{Framework} (b) presents more details regarding the designed transformer blocks. We modify the standard ViT blocks by integrating the designed MoE Adapter layer and MoE LoRA layer with the original blocks. 

Fig.~\ref{Framework} (c) details the designed MoE Adapter layer. During the training process, the MLP parameters are frozen with the ImageNet weights. Within this layer, an MoE module is designed, comprising one gate $G_{A}(\cdot)$ and $M$ forgery Adapter experts. The gating mechanism aims to dynamically select the appropriate experts for each input query, while the designed Adapter experts aim to extract specific local forgery features from the input faces. Fig.~\ref{Framework} (d) illustrates the structure of Adapter $j$. We first reshape the input feature and reduce its channel dimension by using a 1$\times$1 convolution. Subsequently, we design convolution operations to extract specific local forgery clues. The output feature is then passed through to another 1$\times$1 convolution layer to restore the channel dimension. Finally, the feature map is flattened to the original shape. 

Similarly, the attention weights $W_{q,k,v}$ in the MoE LoRA layer are fixed to the ImageNet weights. As shown in Fig.~\ref{Framework} (e), the MoE LoRA module consists of one gate $G_{L}(\cdot)$ and $N$ experts, each with a unique rank. The gate $G_{L}(\cdot)$ selectively activate sparse experts for detecting face forgeries. Fig.~\ref{Framework} (f) depicts the LoRA structure, where $\otimes$ indicates matrix multiplication. The output feature is derived by multiplying the input with two learnable low-rank matrices. The resulting $x_{out}$ is the element-wise summation of the fixed weights and the learned MoE LoRA weights, which are further processed by a self-attention mechanism. \textcolor{black}{Overall, the MoE process can be summarized as follows: 1) The input feature $x_{in}$ is passed into the MoE module; 2) A gating network processes the input and outputs a set of logits, one for each expert; 3) Top-k (k=1 in this work) gating selects the most relevant expert; 4) The input is then routed to the selected expert and output $x_{out}$.}

The whole framework is trained in an end-to-end manner with the supervision of the following objective function:

\begin{equation}
    L = L_{ce} + \lambda \cdot L_{moe},
\end{equation}
where $L_{ce}$ represents the cross-entropy loss. Following \cite{shazeer2017outrageously, bengio2015conditional}, we apply an additional loss $L_{moe}$ to encourage all experts to have equal importance. Further details on $L_{moe}$ will be dedicated at the end of this section.

\subsection{Details of the MoE}
\subsubsection{MoE LoRA Layer.} LoRA modules with different ranks tend to project the input tokens into various feature spaces. However, in uncontrolled deployment environments, it is challenging to manually predefine an ideal rank for different testing faces. To tackle this challenge, we design an MoE LoRA Layer that learns an optimal LoRA expert for each input query.  
As shown in Fig.~\ref{Framework} (e), each LoRA expert $E_{L}(\cdot)_{i}$ specifies a rank $r_{i}$. In this work, we design a gating mechanism $G_{L}(\cdot)$ to dynamically select the Top-k experts (default: k=1). The details of the gating mechanism will be elaborated later. As such, the output tokens can be calculated by:
\begin{equation}
     x_{out} = W_{q/k/v}x_{in} + \sum_{i}^{N} G_{L}(x_{in})_{i} E_{L}(x_{in})_{i},
\end{equation}
where $W_{q/k/v}$ is fixed as the ImageNet weight during the training process. Each LoRA expert is formulated by: 
\begin{equation}
     E_{L}(x_{in})_{i} =x_{in}W_{i}^{down}W_{i}^{up},
\end{equation}
where $W_{i}^{up} \in \mathbb{R}^{r_{i}\times dim}$ and $W_{i}^{down} \in \mathbb{R}^{dim\times r_{i}}$ are two trainable matrices.

\begin{table*}[]
  \caption{\textcolor{black}{Cross-dataset evaluation on five unseen datasets. `*' indicates the trained model provided by the authors. `\dag' indicates our re-implementation using the public official code. Methods in the bottom table denote video-level results. \#Params indicates the activated parameter number during training.}}
  \label{cross-dataset2}
\centering
\scalebox{1.0}{\begin{tabular}{ccccccccccccccc}
\hline
\multirow{2}{*}{Method} & \multirow{2}{*}{Venue} & \multirow{2}{*}{\#Params} & \multicolumn{2}{c}{CDF} & \multicolumn{2}{c}{WDF} & \multicolumn{2}{c}{DFDC-P} & \multicolumn{2}{c}{DFD} & \multicolumn{2}{c}{DFR} & \multicolumn{2}{c}{AVG} \\ \cline{4-15} 
& &  & AUC & EER & AUC & EER & AUC & EER & AUC & EER & AUC & EER & AUC & EER\\ \hline
Face X-ray \cite{li2020face} & CVPR20 & 41.97M & 74.20 & - & - & - & 70.00 & - & 85.60 & - & - & - & - & - \\
GFF \cite{luo2021generalizing} & CVPR21 & 53.25M & 75.31 & 32.48 & 66.51 & 41.52 & 71.58 & 34.77 & 85.51 & 25.64 & - & - & - & - \\
LTW \cite{sun2021domain} & AAAI21 & 20.37M & 77.14 & 29.34 & 67.12 & 39.22 & 74.58 & 33.81 & 88.56 & 20.57  & - & - & - & - \\
F2Trans-S \cite{miao2023f} & TIFS23 & 117.52M& 80.72 & - & - & - & 71.71 & - & -  & - & -  & - & - & -\\
ViT-B \cite{dosovitskiy2020image} & ICLR21 & 85.80M & 72.35 & 34.50 & 75.29 & 33.40 & 75.58 & 32.11 & 79.61 & 28.85 & 80.47 & 26.97 & 76.66 & 31.17 \\
SBI* \cite{shiohara2022detecting} & CVPR22 & 19.34M & 81.33 & 26.94 & 67.22 & 38.85 & \underline{79.87} & \underline{28.26} & 77.37 & 30.18 & 84.90 & 23.13 & 78.14 & 29.47 \\
DCL* \cite{sun2022dual} & AAAI22 & 19.35M & 81.05 & 26.76 & 72.95 & 35.73  & 71.49 & 35.90 & 89.20 & 19.46 & 92.26 & 14.81  & 81.39 & 26.53 \\
Xception\dag \cite{chollet2017xception} & ICCV19 & 20.81M & 64.14 & 39.77 & 68.90 & 38.67 & 69.56 & 36.94 & 84.31 & 25.00 & 91.93 & 15.52 & 75.77 & 31.18 \\
RECCE\dag \cite{cao2022end} & CVPR22 &  25.83M & 61.42 & 41.71 & 74.38 & 32.64 & 64.08 & 40.04 & 83.35 & 24.57 & 92.93 & 14.74 & 75.23 & 30.73 \\
\textcolor{black}{SFDG} \cite{wang2023dynamic}& CVPR23 & 22.40M & 75.83 & 30.30 & 69.27 & 37.70 & - & - & \textbf{92.10} & \textbf{15.10} & - & - & - & - \\
\textcolor{black}{UCF*} \cite{yan2023ucf} & CVPR23 & 46.80M & 77.15& 30.20 & 77.40  & 29.80 & 69.30 & 35.80 & 80.80 & 27.00 & 68.30 & 36.70 & 73.56 & 32.88 \\
\textcolor{black}{LSDA} \cite{yan2024transcending} & CVPR24 & 20.81M & 82.70 & - & 71.40 & - & 76.30 & - & 87.50 & - & 84.20 & - & 78.87 & - \\
EN-B4\dag \cite{tan2019efficientnet} & ICML19 & 19.34M & 65.24 & 39.41 & 67.89 & 37.21 & 67.96 & 37.60 & 88.67 & 18.46 & 92.18 & 15.51 & 76.39 & 29.64 \\
CFM* \cite{10315169} & TIFS24 &  25.37M & \underline{82.78} & \underline{24.74} & \underline{78.39} & \underline{30.79} & 75.82 & 31.67 & \underline{91.47} & \underline{16.80} & {\underline{95.18}} & {\underline{11.87}} & \underline{83.94} & \underline{24.20} \\ 	
MoE-FFD & Ours & \textbf{15.51M} & \textbf{86.69} & \textbf{22.06} & \textbf{80.64} & \textbf{27.11} & \textbf{80.83} & \textbf{26.67} & {90.37} & {17.67} & \textbf{95.39} & \textbf{11.29} & \textbf{86.78} & \textbf{20.96} \\
\hline
\color{black}{Lisiam} \cite{wang2022lisiam} & TIFS22 & - & 78.21 & - & - & - & - & - & - & - & - & - & - & - \\
\color{black}{F$^{3}$Net} \cite{qian2020thinking} & ECCV20 & 42.53M & 68.69 & - & - & - & 67.45 & - & - & - & - & - & - & - \\
\color{black}{FTCN} \cite{zheng2021exploring} & ICCV21 & 26.60M & 86.90 & - & - & - & 74.00 & - & - & - & - & - & - & - \\
\color{black}{ViT-B} \cite{dosovitskiy2020image} & ICLR21 & 85.80M & 78.12 & 30.59 & 78.95 & 29.59 & 79.43 & 28.62 & 84.52 & 23.48 & 86.42 & 19.90 & 81.49 & 26.44 \\ 	
\color{black}{SBI}* \cite{shiohara2022detecting} & CVPR22 & 19.34M & 88.61 & 19.41 & 70.27 & 37.63 & \underline{84.80} & \underline{25.00} & 82.68 & 26.72 & 90.04 & 17.53 & 83.28 & 25.26 \\
\color{black}{DCL}* \cite{sun2022dual} & AAAI22 & 19.35M & 88.24 & 19.12 & 76.87 & 31.44 & 77.57 & 29.55  & \underline{93.91} & 14.40 & 97.41 & 9.96 & 86.80 & 20.89 \\
\color{black}{RECCE}\dag \cite{cao2022end} & CVPR22 & 25.83M & 69.25 & 34.38 & 76.99 & 30.49 & 66.90 & 39.39 & 86.87 & 21.55 & 97.15 & 9.29 & 79.42 & 27.01 \\
\textcolor{black}{UCF*} \cite{yan2023ucf} & CVPR23 & 46.80M & 83.70  & 28.20 & 77.20  & 26.10 & 71.00  & 33.50 &  86.40 & 20.40 & 69.70  & 34.20 & 76.48 & 29.70 \\
\textcolor{black}{LSDA} \cite{yan2024transcending} & CVPR24 & 20.81M & \underline{89.80} & - & 75.60 & - & {81.20} & - & 89.20 & - & 84.20 & - & 82.25 & - \\
\color{black}{CFM}* \cite{10315169} & TIFS24 & 25.37M& {89.65} & \underline{17.65} & \underline{82.27} & \underline{26.80} & 80.22 & 27.48 & \textbf{95.21} & \textbf{11.98} & \underline{97.59} & \underline{9.04} & \underline{88.99} & \underline{18.59} \\
\color{black}{MoE-FFD} & Ours & \textbf{15.51M} & \textbf{91.28} & \textbf{17.15} & \textbf{83.91} & \textbf{24.75} & \textbf{84.97} & \textbf{23.44} & {93.57} & \underline{14.05} & \textbf{98.52} & \textbf{5.47} & \textbf{90.45} & \textbf{16.97} \\
\hline
\end{tabular}} 										
\end{table*}

\subsubsection{MoE Adapter Layer.} Previous studies \cite{shao2022detecting, fei2022learning, yang2021mtd, yu2020searching, su2021pixel} have demonstrated the effectiveness of utilizing local difference convolution to capture face forgery clues. However, these CNN-based methods only limit their scopes in one specific forgery feature. In our research, we introduce the MoE Adapter Layer integrated into the ViT backbone. This design aims to scale the model capacity and facilitate the dynamic selection of the suitable forgery Adapter expert. Additionally, the designed MoE Adapter layer dynamically injects local forgery priors into the plain ViT backbone. 

As shown in Fig.~\ref{Framework} (c), the designed MoE Adapter layer consists of $M$ Convpass Adapter experts. And the output $x_{out}$ can be formularized as:
\begin{equation}
     x_{out} = {\rm MLP}(x_{in}) + \sum_{j}^{M} G_{A}(x_{in})_{j} E_{A}(x_{in})_{j},
\end{equation}
where MLP is frozen as the ImageNet weights during training. Adapter expert $E_{A}(\cdot)_{j}$ can be calculated by (we omit the activation layer):
\begin{equation}
     E_{A}(x_{in})_{j} = {\rm Conv_{1\times 1}^{up}}({\rm Conv_{3\times 3}^{j}}({\rm Conv_{1\times 1}^{down}}(x_{in}))),
\end{equation}
$\rm Conv_{1\times 1}^{down}$ and $\rm  Conv_{1\times 1}^{up}$ are two $1\times 1$ convolution layers, which down-sample and up-sample the channels, respectively. $\rm Conv_{3\times 3}^{j}$ indicates the specific convolution layer in different experts. We develop $M=5$ types of convolution. These convolutions aim to model different local interactions, facilitating the model to capture abundant local forgery features from input faces.

Fig.~\ref{MoE_adapter} shows the designed five convolutions, including (a) Vanilla Convolution, (b) Angular Difference Convolution (ADC), (c) Central Difference Convolution (CDC), (d) Radial Difference Convolution (RDC), and (e) Second-Order Convolution (SOC). The red arrow \textcolor{purple}{$\rightarrow$} indicates the subtraction operation. We first formulate the vanilla convolution as (we omit the bias for concision):
\begin{equation}
     y = \sum_{p\in \Omega_{1}} w_{p}x_{p},
\end{equation}
then, for the other four convolution types, Eq.(10) can be rewritten as:
\begin{equation}
     y = \sum_{p\in \Omega_{j}} w_{p} \hat{x}_{p}= w_{c}{x}_{c} + \sum_{p\in \Omega_{j}, p\ne c} w_{p} \hat{x}_{p},  
\end{equation}
where ${x}_{c}$ and ${w}_{c}$ represent the center elements of the input $x$ and weight $w$. 

In ADC and CDC, we set the $\Omega$ size as 3$\times$3. The $\hat{x}_{p}$ in CDC can be calculated by: $\hat{x}_{p}=x_{p}-x_{c}$. And the $\hat{x}_{p}$ in ADC is denoted as: $\hat{x}_{p}={x}_{p}-{x}_{p}^{next}$, where ${x}_{p}^{next}$ is the next element in the clockwise direction.

In the RDC and SOC operations, we set the size of the  $\Omega$ region as 5$\times$5. For each element ${x}_{p}$, we define a corresponding radial element ${x}_{p}^{R}$ in the peripheral region of $\Omega$, which is highlighted in dark green in Fig.~\ref{MoE_adapter} (d) and (e). As such, the $\hat{x}_{p}$ in RDC is calculated by: $\hat{x}_{p}={x}_{p}^{R}-{x}_{p}$. SOC aims at learning second-order local anomaly of the input. As such, $\hat{x}_{p}$ in SOC is formulated as: $\hat{x}_{p}=(x_{p}^{R}-x_{p})-(x_{p}-x_{c})=(x_{p}^{R}-x_{p})+(x_{c}-x_{p})$. As such, the designed MoE Adapter layer can effectively search intrinsic detailed local forgery patterns in a larger feature space. 

\subsubsection{Gating Network.} 
\textcolor{black}{The gating network in a Mixture-of-Experts (MoE) model dynamically routes each input to the most appropriate expert, determining the contribution of each expert to the final prediction. It performs dynamic routing, enabling the MoE model to adaptively select appropriate experts that are most relevant for each input sample.}
We adopt Top-k noisy gating \cite{shazeer2017outrageously} as our gating mechanism. The gating scores $G(x) \in \mathbb{R}^{N_{e}}$ are determined by the values $H(x) \in \mathbb{R}^{N_{e}}$, where $N_{e}$ indicates the expert number. For a given input $x \in \mathbb{R}^{N_{t} \times dim}$, we first apply average pooling and reshape it to $x_{m} \in \mathbb{R}^{dim}$. Then, we calculate $H(x)$ by: 
\begin{equation}
     H(x) = x_{m} \otimes W_{gate} + {\rm StandardNormal()}({\rm Softplus}(x_{m} \otimes W_{noise})),
\end{equation}
where $W_{gate} \in \mathbb{R}^{dim \times N_{e}}$ and $W_{noise} \in \mathbb{R}^{dim \times N_{e}}$ are slim trainable parameters. $\rm Softplus$ is an activation function. For a given $H(x) \in \mathbb{R}^{N_{e}}$, we keep Top-k ($k\leq N_{e}$) values while setting others as $-\infty$. Then, the gating scores $G(x)$ can be calculated by:
\begin{equation}
     G(x) = {\rm Softmax}({\rm Topk}(H(x),k)).
\end{equation}
To prevent the gating network from converging to a state where it produces large weights for the same few experts. We further apply a soft constraint on the batch-wise average of each gate. As such, for a given batch of data $X$, the MoE loss $L_{moe}$ is defined as:
\begin{equation}
     L_{moe} = CV(Importance(X))^{2};  ~Importance(X) = \sum_{x\in X} G(x),
\end{equation}
where $CV(\cdot)$ indicates the Coefficient of Variation:
\begin{equation}
     CV(Importance(X)) = \frac{{\rm Std}[Importance(X)]}{{\rm Mean}[Importance(X)]}.
\end{equation}
\textcolor{black}{As such, the additional loss $L_{moe}$ encourages all experts to have equal importance and allows the model to avoid relying on a single expert during training, which can effectively prevent overfitting in face forgery detection.}

\section{Experiments}
In this section, we evaluate our model in terms of generalizability, robustness, and parameter-efficiency under a wide variety of experimental settings. We conduct extensive ablation studies to demonstrate the effectiveness of the designed network architecture and the adopted training strategy.

\subsection{Implementation Details}
We apply the popular MTCNN face detector \cite{zhang2016joint} to crop the face regions. The proposed framework is implemented on the Pytorch \cite{paszke2019pytorch} platform. The model is trained using Adam optimizer \cite{kingma2014adam} with $\beta_{1}$ = 0.9 and $\beta_{2}$ = 0.999. We set the learning rate of gating network and other trainable parameters as 1e-4 and 3e-5, respectively. The loss weight $\lambda$ in Eq. (5) is set as 1. We train the model for 20 epochs on one single 3090 GPU with batch size 32. We follow the official dataset split strategy in \cite{10315169} for fair comparison. \textcolor{black}{In the proposed MoE-FFD framework, the ViT-Base model pre-trained on ImageNet-21K \cite{ridnik2021imagenet} serves as the backbone. It consists of 12 transformer blocks, each comprising a 12-head self-attention layer and a multi-layer perceptron (MLP) layer.}


\begin{table}
  \textcolor{black}{\caption{Cross-dataset evaluation on DF40 (CDF).}
  \label{df40-data}
  \centering
  \renewcommand\arraystretch{1.15}
  \scalebox{0.88}{\begin{tabular}{ccccccccc}
\hline
\multirow{2}{*}{Method}  & \multicolumn{2}{c}{FS (CDF)} & \multicolumn{2}{c}{FR (CDF)} & \multicolumn{2}{c}{EFS (CDF)} & \multicolumn{2}{c}{AVG}\\ \cline{2-9} 
 & AUC & EER & AUC & EER & AUC & EER & AUC & EER\\ \hline
SBI\cite{shiohara2022detecting}  & 70.01 & 35.20 & 48.90 & 50.98 & 65.61 & 39.21 & 61.50 & 41.79\\ \hline			
CFM\cite{10315169}   & \underline{78.65} & \underline{28.73} & \textbf{69.11} & \textbf{35.96} & \underline{71.49} & \underline{34.33} & \underline{73.08} & \underline{33.00}\\ \hline
DCL\cite{sun2022dual}   & 72.67 & 35.04 & 62.88 & 41.01 & 63.18 & 41.20 & 66.24 & 39.08\\ \hline
ViT-B \cite{dosovitskiy2020image} & 67.78 & 37.19 & 55.65 & 46.57 & 60.50 & 42.64 & 61.31 & 42.13\\ \hline
MoE-FFD & \textbf{82.11} & \textbf{26.97} & \underline{66.54} & \underline{37.92} & \textbf{78.17} & \textbf{29.78} & \textbf{75.60} & \textbf{31.55}\\ \hline
\end{tabular}}}
\end{table}

\begin{table}
  \caption{Cross-manipulation detection AUC on the unseen manipulation technique (FF++ C23 dataset).}
  \label{cross-mani1}
  \centering
  \renewcommand\arraystretch{1.15}
  \scalebox{1.0}{\begin{tabular}{ccccccc}
\hline
   {Method} & {\#Params} & DF & FF & FS & NT & {AVG}\\
    \hline			
    Xception \cite{chollet2017xception} & 20.81M & 0.907 & 0.753 & 0.460 & 0.744 & 0.716 \\ 		
    \hline	 				
    EN-B4 \cite{tan2019efficientnet}  & 19.34M & 0.485 & 0.556 & 0.517 & 0.493 & 0.513 \\ 
    \hline	 				
    AT EN-B4 \cite{tan2019efficientnet} & 19.34M & 0.911 & 0.801 & 0.543 & 0.774 & 0.757 \\	 
    \hline	 				
    FL EN-B4 \cite{tan2019efficientnet} & 19.34M & 0.903 & 0.798 & 0.503 & 0.759 & 0.741 \\ 
    \hline						
    MLDG \cite{li2018learning} & 62.38M & 0.918 & 0.771 & 0.609 & \textbf{0.780} & 0.770 \\ 
    \hline	 								
    LTW \cite{sun2021domain} & 20.37M & \underline{0.927} &  0.802 & 0.640 & 0.773 & \underline{0.786} \\ 
    \hline	
    ViT-B \cite{dosovitskiy2020image} & 85.80M & 0.771 & 0.656 & 0.510 & 0.554 & 0.623 \\		
    \hline						
    CFM \cite{10315169} & 25.37M & 0.880 & \underline{0.814} & \underline{0.630} & 0.643& 0.742  \\
    \hline
    MoE-FFD & \textbf{15.51M} & \textbf{0.947} & \textbf{0.877} & \textbf{0.647} & \underline{0.759} & \textbf{0.808} \\
    \hline
\end{tabular}}
\end{table}

\subsection{Datasets and Evaluation Metrics}

The experiments encompass the following six datasets for training and testing: FaceForensics++ (FF++) \cite{rossler2019faceforensics++}, CelebDF-v2 (CDF) \cite{li2020celeb}, WildDeepfake (WDF) \cite{zi2020wilddeepfake}, DeepFake Detection Challenge Preview (DFDC-P) \cite{dolhansky2019deepfake}, DeepFakeDetection (DFD) \cite{dfd.org}, and DeepForensics-1.0 (DFR) \cite{jiang2020deeperforensics}. FF++ is a widely used dataset in face forgery detection, which includes four face manipulation types: Deepfakes (DF) \cite{DeepFake}, Face2Face (FF) \cite{thies2016face2face}, FaceSwap (FS) \cite{FaceSwap}, and NeuralTextures (NT) \cite{thies2019deferred}. 
The remaining datasets represent recent five Deepfake datasets with diverse environment variables and various video qualities. 
We perform cross-dataset and cross-manipulation evaluations to examine the models' generalization capability. \textcolor{black}{Specifically, CDF is a high-quality DeepFake dataset specifically focused on celebrity identities. WDF is a web-crawled DeepFake dataset that contains videos with diverse scenes and forgery methods. DFDC-P is a challenging dataset that contains deepfake videos generated from various scenarios using two face manipulation methods. DFD is a comprehensive DeepFake dataset comprising over 3,000 manipulated videos involving 28 actors across various scenes. DFR is created by using real videos from FF++ with an innovative end-to-end face swapping framework.}
\textcolor{black}{In addition, we examine our method's effectiveness on the recently released DF40 benchmark \cite{yan2024df40} with 40 forgery types. It comprises 10 face swapping (FS) methods, 13 face reenactment (FR) methods, 12 entire face synthesis (EFS) methods, and 5 face editing (FE) methods. Real data is sourced from the FF++ \cite{rossler2019faceforensics++} and CDF \cite{li2020celeb} datasets, and the corresponding fake data is generated to construct domain-specific subsets for each manipulation type, such as FS (FF++) and FS (CDF). }
Furthermore, we measure the model's robustness by applying it to perturbed face images with various perturbation types and severity levels. Consistent with prior arts, we adopt Area under the ROC Curve (AUC) and Equal Error Rate (EER) as evaluation metrics. AUC measures the Area under the Receiver Operating Characteristic (ROC) curve, while EER denotes the False Positive Rate (FPR) that equals to the True Positive Rate (TPR).

\begin{table}
  \caption{Cross-manipulation detection AUC on the remaining three unseen manipulation techniques (FF++ C23 dataset). }
  \label{cross-mani2}
  \centering
  \renewcommand\arraystretch{1.15}
  \scalebox{1.06}{\begin{tabular}{c|c|ccccc} 
\hline
Methods & Train & DF & FF & FS & NT & AVG$^{*}$ \\ \hline
ViT \cite{dosovitskiy2020image} & & \cellcolor[HTML]{E0DBDB}99.28 & 59.87 & 49.91 & 62.38 & 57.39\\ 
RECCE \cite{cao2022end} & & \cellcolor[HTML]{E0DBDB}99.95 & 69.75 & \textbf{54.72} & 77.15 & 67.21 \\
MoE-FFD & \multirow{-3}{*}{DF} & \cellcolor[HTML]{E0DBDB}99.80 & \textbf{73.46} & 52.15 & \textbf{77.45} & \textbf{67.69} \\ \hline
ViT \cite{dosovitskiy2020image} & & 74.72 & \cellcolor[HTML]{E0DBDB}99.21 & 57.19 & 56.38 & 62.76\\ 
RECCE \cite{cao2022end} & & 71.55 & \cellcolor[HTML]{E0DBDB}99.20 & 50.02 & \textbf{72.27} & 64.61 \\
MoE-FFD & \multirow{-3}{*}{FF} & \textbf{86.67} & \cellcolor[HTML]{E0DBDB}99.43 & \textbf{66.92} & 68.75 & \textbf{74.11}\\ \hline
ViT \cite{dosovitskiy2020image} & & 78.59 & 61.62 & \cellcolor[HTML]{E0DBDB}99.44 & 46.69 & 62.30 \\ 
RECCE \cite{cao2022end} & & 63.05 & 66.21 & \cellcolor[HTML]{E0DBDB}99.72 & \textbf{58.07} & 62.44 \\
MoE-FFD & \multirow{-3}{*}{FS} & \textbf{79.89} & \textbf{71.45} & \cellcolor[HTML]{E0DBDB}99.56 & 48.33 &\textbf{66.56}\\ \hline
ViT \cite{dosovitskiy2020image} & & 78.46 & 68.31 & 45.07 & \cellcolor[HTML]{E0DBDB}97.19 & 63.95\\ 
RECCE \cite{cao2022end} & & 72.37 & 64.69 & 51.61 & \cellcolor[HTML]{E0DBDB}99.59 & 62.89 \\
MoE-FFD & \multirow{-3}{*}{NT} & \textbf{80.02} & \textbf{73.02} & \textbf{51.94} & \cellcolor[HTML]{E0DBDB}98.70 & \textbf{68.33}\\ \hline
\end{tabular}}
\end{table}

\begin{table}[]
\centering
\textcolor{black}{\caption{Cross-manipulation evaluation on DF40 (FF++). (AUC)} 
\label{df40-mani}
\scalebox{0.85}{\begin{tabular}{c|c|cccc}
\hline
\multirow{2}{*}{Training Set} & \multirow{2}{*}{Model} & \multicolumn{4}{c}{Testing Set} \\ \cline{3-6} 
 & & FS (FF++) & FR (FF++) & EFS (FF++) & AVG (FF++) \\
\hline
\multirow{8}{*}{FS (FF++)} 
& Xception \cite{chollet2017xception} & \cellcolor[HTML]{E0DBDB}0.991 & 0.892 & 0.810 & 0.898 \\
& CLIP \cite{radford2021learning} & \cellcolor[HTML]{E0DBDB}\textbf{0.996} & \underline{0.908} & \underline{0.837} & \underline{0.914} \\
& SRM \cite{luo2021generalizing} & \cellcolor[HTML]{E0DBDB}0.988 & 0.867 & 0.703 & 0.853 \\
& SPSL \cite{liu2021spatial}  & \cellcolor[HTML]{E0DBDB}0.987 & 0.849 & 0.735 & 0.857 \\
& RECCE \cite{cao2022end} & \cellcolor[HTML]{E0DBDB}0.991 & 0.855 & 0.758 & 0.868 \\
& RFM \cite{wang2021representative}  & \cellcolor[HTML]{E0DBDB}\underline{0.992} & 0.884 & 0.821 & 0.899 \\
& ViT-B \cite{dosovitskiy2020image}  & \cellcolor[HTML]{E0DBDB}0.983 & 0.877 & 0.771 & 0.877 \\
& MoE-FFD &\cellcolor[HTML]{E0DBDB}0.990  & \textbf{0.979}  & \textbf{0.904}  &\textbf{0.957} \\
\hline
\multirow{8}{*}{FR (FF++)} 
& Xception \cite{chollet2017xception} & 0.838 & \cellcolor[HTML]{E0DBDB}0.996 & 0.670 & 0.835 \\
& CLIP \cite{radford2021learning} & \underline{0.932} & \cellcolor[HTML]{E0DBDB}\textbf{0.999} & \underline{0.798} & \underline{0.910} \\
& SRM \cite{luo2021generalizing} & 0.893 & \cellcolor[HTML]{E0DBDB}\underline{0.998} & 0.698 & 0.863 \\
& SPSL \cite{liu2021spatial} & 0.901 & \cellcolor[HTML]{E0DBDB}\underline{0.998} & 0.695 & 0.865 \\
& RECCE \cite{cao2022end} & 0.865 & \cellcolor[HTML]{E0DBDB}0.997 & 0.716 & 0.859 \\
& RFM \cite{wang2021representative} & 0.892 & \cellcolor[HTML]{E0DBDB}\textbf{0.999} & 0.776 & 0.889 \\
& ViT-B \cite{dosovitskiy2020image}  & 0.764 & \cellcolor[HTML]{E0DBDB}0.986 & 0.645 & 0.798 \\
& MoE-FFD &\textbf{0.933}  &\cellcolor[HTML]{E0DBDB}\textbf{0.999}  &\textbf{0.866}  & \textbf{0.932} \\
\hline
\multirow{8}{*}{EFS (FF++)} 
& Xception  \cite{chollet2017xception} & 0.665 & 0.807 & \cellcolor[HTML]{E0DBDB}\textbf{0.999} & 0.824 \\
& CLIP \cite{radford2021learning} & 0.688 & \textbf{0.889} & \cellcolor[HTML]{E0DBDB}\textbf{0.999} & \underline{0.859} \\
& SRM \cite{luo2021generalizing} & 0.596 & 0.776 & \cellcolor[HTML]{E0DBDB}\textbf{0.999} & 0.790 \\
& SPSL \cite{liu2021spatial} & 0.659 & 0.811 & \cellcolor[HTML]{E0DBDB}\textbf{0.999} & 0.823 \\
& RECCE \cite{cao2022end} & \underline{0.691} & 0.801 & \cellcolor[HTML]{E0DBDB}\textbf{0.999} & 0.830 \\
& RFM \cite{wang2021representative} & 0.653 &0.795 & \cellcolor[HTML]{E0DBDB}\textbf{0.999} & 0.816 \\
& ViT-B \cite{dosovitskiy2020image}  & 0.550 &0.718 & \cellcolor[HTML]{E0DBDB}\underline{0.998} & 0.755 \\
& MoE-FFD & \textbf{0.737}  & \underline{0.854}  &\cellcolor[HTML]{E0DBDB}\textbf{0.999}  &\textbf{0.863} \\
\hline
\end{tabular}}}
\end{table}

\subsection{Comparison with State-of-the-Arts}
\vspace{-7mm}
\textcolor{black}{\subsubsection{Baseline Models.}} 
\textcolor{black}{This paper incorporates 14 representative baseline detectors from top journals and conferences, including three data-driven architectures and 11 state-of-the-art image forgery detectors:}

\noindent \textcolor{black}{\textbf{Face X-ray} \cite{li2020face} detects face forgeries by revealing blending boundaries indicative of image composition from different sources.}

\noindent \textcolor{black}{\textbf{GFF} \cite{luo2021generalizing} leverages high-frequency noise information for face forgery detection, which is less affected by manipulation-specific textures and more indicative of tampered regions.} 

\noindent \textcolor{black}{\textbf{LTW} \cite{sun2021domain} uses meta-weight learning to assign adaptive weights to training samples from different domains, mitigating domain-specific biases and enhancing model generalizability.} 

\noindent \textcolor{black}{\textbf{F2Trans-S} \cite{miao2023f} is a high-frequency fine-grained transformer network for face forgery detection that enhances generalization and robustness by leveraging fine-grained manipulation traces in both spatial and frequency domains.} 

\noindent \textcolor{black}{\textbf{SBI} \cite{shiohara2022detecting}  encourages the model to learn more general and manipulation-agnostic features by blending pseudo source and target images from single pristine images.}

\noindent \textcolor{black}{\textbf{DCL} \cite{sun2022dual} enhances generalization by constructing positive and negative pairs for contrastive learning at both inter- and intra-instance levels.} 

\noindent \textcolor{black}{\textbf{RECCE} \cite{cao2022end} focuses on learning compact and generalizable representations of genuine faces rather than specific forgery patterns. By combining multi-scale reconstruction learning with classification learning, RECCE effectively enhances the model robustness and generalization.}

\noindent \textcolor{black}{\textbf{SFDG} \cite{wang2023dynamic} enhances model's generalization capability by modeling relation-aware features across spatial and frequency domains.}

\noindent \textcolor{black}{\textbf{UCF} \cite{yan2023ucf} is a disentanglement-based framework for deepfake detection that improves generalization by isolating and leveraging common forgery features while minimizing reliance on forgery-irrelevant or method-specific patterns.} 

\noindent \textcolor{black}{\textbf{LSDA} \cite{yan2024transcending} mitigates overfitting to method-specific artifacts and enables a more robust decision boundary by simulating intra- and inter-forgery variations, thereby achieving superior performance across diverse datasets.}

\noindent \textcolor{black}{\textbf{CFM} \cite{10315169} enhances face forgery detection by mining critical forgery features through prior knowledge-agnostic data augmentation, fine-grained relation learning, and a progressive learning controller.} 

\subsubsection{Cross-Dataset Evaluations.}
In this subsection, we conduct cross-dataset experiments to evaluate the models' generalization capability. In practical scenarios, the trained models are vulnerable to unseen domains, leading to significant performance degradations. We train all face forgery detectors on the FF++ (C23) dataset and directly test them on five unseen Deepfake datasets: CDF, WDF, DFDC-P, DFD, and DFR. Table~\ref{cross-dataset2} presents the frame-level and video-level detection results, divided into top and bottom sections. Video-level detection results presented in the bottom table are obtained by averaging all frame scores within each video. We bold the best results and underline the second-best results. Notably, MoE-FFD achieves the highest detection performance on four datasets in both frame-level and video-level detection. Additionally, MoE-FFD's performance is on par with CFM on the DFD dataset. Moreover, MoE-FFD outperforms CFM by a clear margin in average performance. The average frame-level AUC improves by 2.86\%, rising from 83.94\% to 86.78\%. Similarly, MoE-FFD also outperforms others in video-level detection on average. Our MoE-FFD's generalizability improvements over existing methods offers valuable insights into the effectiveness of the proposed approach. 

\textcolor{black}{We present the cross-dataset evaluation results on the DF40 dataset in Table~\ref{df40-data}. The listed models are trained on FF++ (C23) and evaluated on FS (CDF), FR (CDF), and EFS (CDF), which are generated from the CDF dataset. It can be observed that MoE-FFD exhibits superior generalization performance on the unforeseen datasets, further validating its effectiveness for face forgery detection.}

\textcolor{black}{Unlike previous arts, MoE-FFD adopts a parameter-efficient fine-tuning strategy that preserves pretrained knowledge while dynamically adapting to forgery-specific features. Furthermore, MoE-FFD effectively scales up the network’s capacity by introducing multiple specialized forgery experts, enabling the model to dynamically select the most appropriate expert when processing unseen facial forgeries.}

\begin{figure*}[ht]
\centering 
\includegraphics[scale=0.36]{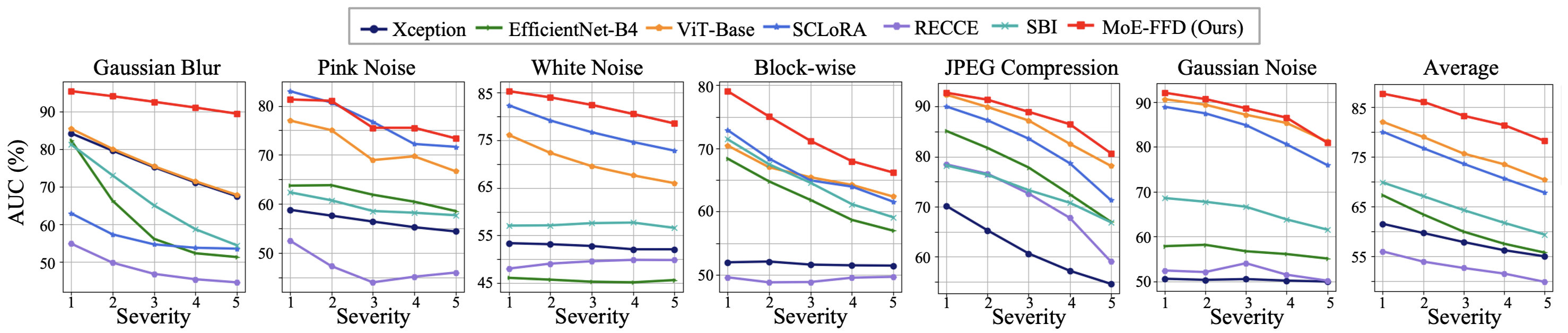}
\caption{\textcolor{black}{Robustness to various common perturbations at five severity levels: Gaussian blur, pink noise, white noise, block wise, JPEG compression, and Gaussian noise.}} 
\label{fig_Robustness}
\end{figure*}

\subsubsection{Cross-Manipulation Evaluations.} 
Existing face forgery detectors often struggle to handle emerging manipulation techniques. With the rapid development of AIGC, more sophisticated manipulation techniques continue to emerge, posing challenges for existing detection models. Ensuring model generalizability to unseen forgeries is crucial for real-world applications. In this study, we conduct cross-manipulation experiments involving four forgery techniques: Deepfakes (DF), Face2Face (FF), FaceSwap (FS), and NeuralTextures (NT). Table~\ref{cross-mani1} presents the results, where models trained on three manipulation types are tested on the remaining one.  MoE-FFD exhibits the best average detection results. Compared to the ViT-B baseline, the proposed method achieves an average AUC enhancement of 18.5\%, going from 62.3\% to 80.8\%. In Table~\ref{cross-mani2}, we further examine models trained on one manipulation type and tested across the other three. AVG$^{*}$ denotes the average AUC score across three cross-manipulation trials. MoE-FFD outperforms the baseline in both intra- and cross-manipulation evaluations. Remarkably, MoE-FFD achieves around 99\% AUC scores in each intra-manipulation evaluation, highlighting its outstanding detection accuracy in intra-domain evaluations. When compared to the state-of-the-art RECCE method, our MoE-FFD consistently achieves superior average cross-manipulation results across all four evaluation settings. \textcolor{black}{In addition, we further perform cross-manipulation evaluations on the DF40 benchmark. Following the official protocol, the model is trained on a single manipulation type (i.e., FS, FR, or EFS) and evaluated across all manipulation types within the FF++ domain. As shown in Table~\ref{df40-mani}, the proposed MoE-FFD consistently achieves the highest average AUC scores across various cross-manipulation settings, demonstrating superior performance compared to other SOTA methods.} These findings suggest that MoE-FFD is generalized to unseen manipulations.

\begin{table}
  \caption{Ablation experiment on the designed MoE modules.}
  \label{PEFT}
  \centering
  \renewcommand\arraystretch{1.15}
  \scalebox{0.88}{\begin{tabular}{ccccccccc}
\hline
\multirow{2}{*}{ViT-B} & {MoE} & {MoE} & \multicolumn{2}{c}{CDF} & \multicolumn{2}{c}{DFDC-P} & \multicolumn{2}{c}{WDF}  \\ \cline{4-9} 
& LoRA & Adapter & AUC & EER & AUC & EER & AUC & EER \\ \hline
\checkmark & - & - & 72.35 & 34.50 & 75.58 & 32.11 & 75.29 & 33.40 \\ \hline				
\checkmark & \checkmark & - & 84.84 & 23.51 & 79.62 & 28.13 & 79.73 & 28.62 \\ \hline
\checkmark & - & \checkmark &  83.21& 25.34 & 76.35 & 30.90 & 77.15 & 30.93 \\ \hline
\checkmark & \checkmark & \checkmark & \textbf{86.69} & \textbf{22.06} & \textbf{80.83} & \textbf{26.67} & \textbf{80.64} & \textbf{27.11} \\ \hline
\end{tabular}}
\end{table}

\subsubsection{Robustness to Real-World Perturbations.} 
Images and videos transmitted online always undergo various perturbations that erase forgery cues within the image/video contents \cite{wu2022robust}. \textcolor{black}{Existing Deepfake detectors always suffer from performance drops on distorted data. In this work, we introduce common perturbations to evaluate the model’s robustness, including Gaussian blurring, pink noise, white noise, block-wise, JPEG compression, and Gaussian noise. Each perturbation type includes five severity levels to mimic diverse real-world conditions. }
Note that we do not apply any data augmentations during training, such that the tested perturbations are totally unseen for our model. As shown in Fig.~\ref{fig_Robustness}, the detectors' AUC detection performance consistently deteriorates with the increasing severity level. \textcolor{black}{Nevertheless, our proposed MoE-FFD demonstrates higher resilience to most perturbations compared to existing methods. Notably, it achieves substantial robustness gains over the ViT-B baseline, demonstrating the effectiveness of the proposed MoE learning strategy and PEFT modules in improving robustness against image perturbations. This further suggests that MoE-FFD can effectively capture intrinsic forgery features.}


\subsubsection{Discussion.} 
MoE-FFD achieves superior generalizability and robustness compared to previous methods, which can be attributed to the following designs: (1) MoE-FFD only updates external modules while preserving the abundant ImageNet knowledge, enabling the model to adaptively learn forgery-specific features; (2) MoE-FFD integrates the LoRA and Convpass Adapter with the ViT backbone, effectively leveraging the expressivity of transformers and the local forgery priors; (3) The incorporation of MoE modules facilitates optimal selection of LoRA and Adapter experts for  forgery feature mining. Furthermore, MoE-FFD presents a parameter-efficient approach to face forgery detection due to the utilized PEFT strategy. To validate the effectiveness of the designed components, we perform ablation experiments next.

\begin{table}
  \vspace{1mm}
  \caption{Cross-dataset detection performance on different ViT backbones.}
  \label{backbone}
  \centering
  \renewcommand\arraystretch{1.15}
  \scalebox{0.8}{\begin{tabular}{cccccccccc}
\hline
\multirow{2}{*}{Backbone} & \multirow{2}{*}{MoE-FFD} & \multirow{2}{*}{\#Params} & \multicolumn{2}{c}{CDF} & \multicolumn{2}{c}{DFDC-P} & \multicolumn{2}{c}{WDF} \\ \cline{4-9}
& & & AUC & EER & AUC & EER & AUC & EER \\ \hline	
ViT-Tiny & - & 5.52M & 66.41 & 38.53 & 71.92 & 34.23 & 69.71 & 37.23 \\
\cellcolor[HTML]{E0DBDB} ViT-Tiny & \cellcolor[HTML]{E0DBDB}\checkmark & \cellcolor[HTML]{E0DBDB}3.90M & \cellcolor[HTML]{E0DBDB}76.56 & \cellcolor[HTML]{E0DBDB}31.13 & \cellcolor[HTML]{E0DBDB}73.61 & \cellcolor[HTML]{E0DBDB}32.61 & \cellcolor[HTML]{E0DBDB}75.40 & \cellcolor[HTML]{E0DBDB}31.50 \\ \hline 
ViT-Small & - & 21.67M & 70.03 & 35.71 & 72.19 & 34.02 & 71.67 & 35.66 \\
\cellcolor[HTML]{E0DBDB}ViT-Small & \cellcolor[HTML]{E0DBDB}\checkmark & \cellcolor[HTML]{E0DBDB}7.73M & \cellcolor[HTML]{E0DBDB}81.22 & \cellcolor[HTML]{E0DBDB}27.58 & \cellcolor[HTML]{E0DBDB}78.08 & \cellcolor[HTML]{E0DBDB}29.22 & \cellcolor[HTML]{E0DBDB}77.83 & \cellcolor[HTML]{E0DBDB}30.23 \\ \hline
ViT-Large & - & 303.30M & 73.13 & 33.28 & 74.46 & 32.25 & 72.96 & 34.45 \\
\cellcolor[HTML]{E0DBDB}ViT-Large & \cellcolor[HTML]{E0DBDB}\checkmark & \cellcolor[HTML]{E0DBDB}41.34M & \cellcolor[HTML]{E0DBDB}86.21 & \cellcolor[HTML]{E0DBDB}22.11 & \cellcolor[HTML]{E0DBDB}77.51 & \cellcolor[HTML]{E0DBDB}29.45 & \cellcolor[HTML]{E0DBDB}80.00 & \cellcolor[HTML]{E0DBDB}28.33 \\ \hline
\end{tabular}}
\end{table}

\subsection{Ablation Experiments}

\subsubsection{Impacts of the PEFT Layers.} 
The proposed MoE-FFD integrates one LoRA layer and one Adapter layer with each ViT block. The LoRA layer is designed to learn forgery-specific parameters for subsequent attention mechanisms, capturing long-range interactions within the input faces. Meanwhile, the Convpass Adapter introduces forgery local priors into the plain ViT model. To study the effectiveness of the designed PEFT modules, we report the cross-dataset detection results in Table~\ref{PEFT}. Utilizing MoE LoRA and MoE Adapter significantly improves the model generalizability across all datasets compared to the vanilla ViT-B backbone. This enhancement can be attributed to the LoRA and Adapter layers effectively retaining the ImageNet knowledge while adaptively learning forgery features. Furthermore, the combination of LoRA and Adapter layers allows the model to capture both long-range interactions and local forgery cues, further enhancing its face forgery detection performance.
 
\subsubsection{Effectiveness on Other Backbones.}
Flexibility is crucial in real-world applications to address the complexities of practical scenarios. Deployment devices with varying computational resources may necessitate different detection models. Therefore, we integrate the proposed MoE-FFD into ViT-Tiny, ViT-Small, and ViT-Large models to assess the flexibility of our approach. 
MoE-FFD can be readily inserted to other vision transformer backbones in a plug-and-play manner. The cross-dataset results are presented in Table~\ref{backbone}. Compared with directly finetuning the vanilla ViT backbones, MoE-FFD significantly reduces the training parameters. As the model size increases, the model generalizability consistently enhances. Additionally, significant performance improvements are observed across different ViT backbones by using MoE-FFD. Specifically, when we assemble MoE-FFD with the ViT-Large backbone, the model achieves 13.08\%, 3.05\%, and 7.04\% AUC boosts on CDF, DFDC-P, and WDF datasets, with only $\sim$13.6\% trainable parameters. This, in turn, demonstrates the flexibility and effectiveness of our method.

\begin{table}
  \vspace{1mm}
  \caption{Impacts of the matrix rank of LoRA layers and the effectiveness of MoE.}
  \label{lora_rank}
  \centering
  \renewcommand\arraystretch{1.15}
  \scalebox{1.1}{\begin{tabular}{cccccccc} 
\hline
\multirow{2}{*}{Setting} & \multicolumn{2}{c}{CDF} & \multicolumn{2}{c}{DFDC-P} & \multicolumn{2}{c}{WDF} \\ \cline{2-7}
& AUC & EER & AUC & EER & AUC & EER \\ \hline						
rank=8 & 82.46& 25.94 &  78.30& 29.24 & 79.35 & 28.77 \\ \hline
rank=16 & 82.16 & 26.01 & 77.85 & 29.74 & 79.15 &  29.37\\ \hline
rank=32 & 81.58 & 26.63 & 78.11 & 29.48 & 79.48 & 29.00 \\ \hline
rank=48 & 83.10  & 25.06  & 79.06 & 29.45 & 79.50 & 28.92 \\ \hline
rank=64 &83.85 & 24.97 & 79.38 & 28.19 &  79.46 & \textbf{27.99}  \\ \hline
rank=96 & 83.50 & 24.86 & 79.53 & 27.69 & 78.28 & 29.14 \\ \hline
rank=128 & 83.47 & 24.52 & 77.36 & 30.48  & 78.93 & 29.89  \\ \hline
MoE & \textbf{84.84} & \textbf{23.51} & \textbf{79.62} & \textbf{28.13} & \textbf{79.73} & 28.62  \\ \hline	
\end{tabular}}
\end{table}

\begin{figure*}[ht]
\centering
\includegraphics[scale=0.38]{ 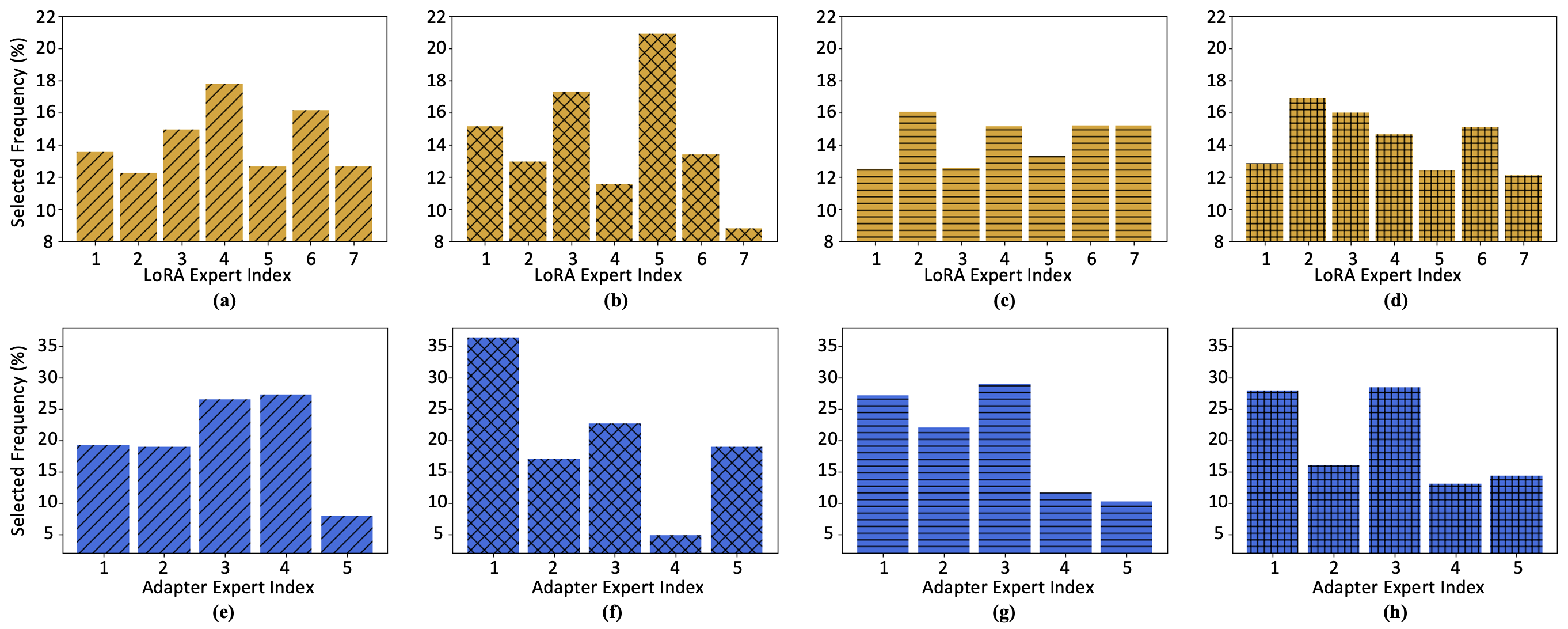}
\caption{LoRA expert selection frequency on (a) CDF,  (b) WDF, (c) DFDC-P, and (d) DFD datasets; Adapter expert selection frequency on (e) CDF, (f) WDF, (g) DFDC-P, and (h) DFD datasets.}
\label{Moe_freq}
\end{figure*}

\subsubsection{Impacts of the MoE Learning Scheme.} 
In this work, the proposed MoE is designed to dynamically select the optimal Top-1 expert for face forgery detection. In Table~\ref{lora_rank}, we compare the cross-dataset detection performance of the proposed MoE-FFD with individual LoRA experts of varying ranks. Notably, different datasets exhibit varying optimal LoRA ranks. The model reaches the best AUC scores on CDF, DFDC-P, and WDF datasets with the LoRA rank of 64, 96, and 48, respectively. MoE dynamically selects the LoRA rank, facilitating the model to search the optimal feature space for each input query. In Table~\ref{lora_rank}, MoE consistently outperforms individual LoRA experts on all three datasets. 
Similarly, we investigate the impact of different Adapters in Table~\ref{adapter}. As introduced in Sec. 3.3.2, the designed adapters tend to expose different local artifacts of input faces. MoE smartly searches the best local feature extractor for each input real/fake face, thereby achieving the superior results compared to using a single Adapter. 
It should be noted that the MoE approach only introduces negligible additional activated parameters, demonstrating that the performance improvements stem from the proposed MoE learning scheme, instead of the model scaling up.

In Fig.~\ref{Moe_freq}, we further investigate the expert selection distributions on four datasets. Fig.~\ref{Moe_freq} (a)-(d) show the LoRA expert selection frequency on CDF, WDF, DFDC-P, and DFD datasets, while Fig.~\ref{Moe_freq} (e)-(h) illustrate the Adapter expert selection frequency. Different Deepfake datasets generally exhibit significant domain gaps. From Fig.~\ref{Moe_freq}, we observe distinct LoRA and Adapter expert selection distributions among the four datasets. Our model dynamically projects the features into suitable space and extracts informative local features tailored to each dataset. 
This observation underscores MoE-FFD's capability to select optimal experts tailored to different data, offering valuable insights into its adaptive nature.

\begin{table}
  \caption{Impacts of different designed Adapters and the effectiveness of MoE.}
  \label{adapter}
  \centering
  \renewcommand\arraystretch{1.15}
  \scalebox{1.1}{\begin{tabular}{cccccccc} 
\hline
\multirow{2}{*}{Setting} & \multicolumn{2}{c}{CDF} & \multicolumn{2}{c}{DFDC-P} & \multicolumn{2}{c}{WDF} \\ \cline{2-7}
& AUC & EER & AUC & EER & AUC & EER \\ \hline	
Conv & 81.25 & 27.36 & 75.01 & 32.39 & 76.28 & 31.15 \\ \hline
ADC & 77.52 & 29.86 & 76.41 & 30.99 & 76.48 & 31.45 \\ \hline
CDC & 82.50 & 26.12 & 75.66 & 31.66 & 76.06 & 31.21 \\ \hline
RDC & 83.05 & 25.39 & 75.12 & 31.90 & 76.16 & 31.86 \\ \hline
SDC & 78.29 & 29.54 & \textbf{76.80} & 30.93 & 76.87 & 31.35 \\ \hline
MoE & \textbf{83.21} & \textbf{25.34} & 76.35 & \textbf{30.90} & \textbf{77.15} & \textbf{30.93}  \\ \hline	
\end{tabular}}
\end{table}

\begin{table}
  \centering
  \textcolor{black}{\caption{Ablation experiment on the selection of Top-k.} 
  \label{top-k}
  \renewcommand\arraystretch{1.15}
  \scalebox{1.0}{\begin{tabular}{ccccccc}
  \hline
  \multirow{2}{*}{Top-k}  & \multicolumn{2}{c}{CDF} & \multicolumn{2}{c}{DFDC-P} & \multicolumn{2}{c}{WDF}  \\ \cline{2-7} 
   & AUC & EER & AUC & EER & AUC & EER \\ \hline
  k=3  & 84.01 & 24.24 & 80.00 & 28.27 & 78.84 & 29.51 \\ \hline				
  k=2  & 85.29 & 23.62 & 79.49 & 28.48 & 79.44 & 28.82 \\ \hline
  k=1 & \textbf{86.69} & \textbf{22.06} & \textbf{80.83} & \textbf{26.67} & \textbf{80.64} & \textbf{27.11} \\ \hline
\end{tabular}}}
\end{table}

\textcolor{black}{\subsubsection{Impacts of the Number of Activated Experts.}}  
\textcolor{black}{To further investigate the impact of the number of activated experts in the Top-k noisy gating mechanism, we conducted an ablation study, as presented in Table~\ref{top-k}. The results show that selecting the Top-1 expert consistently yields the highest AUC scores across all unseen datasets, demonstrating its superior generalization capability. In contrast, activating multiple experts may introduce noise from less confident or less relevant ones. Given that face forgery detection is a binary classification task, incorporating weaker experts with the accurate Top-1 expert can dilute informative signals \cite{shazeer2017outrageously}, thereby deteriorating the detection performance. }

\vspace{5mm}
\subsubsection{MoE v.s. Multi-Experts.} 
While the proposed MoE learning scheme exhibits superior generalizability compared to using a single expert, it is still ambiguous whether the performance boosts stem from the designed MoE or the joint usage of experts. To further demonstrate the dynamic selection of optimal LoRA and Adapter experts by MoE, we compare MoE with Multi-Experts (Multi-E). Multi-E aggregates the features of all designed experts without gating operation. The efficiency and the face forgery detection performance of MoE and Multi-E are reported in Table~\ref{MultiE}. Thanks to the used gating mechanism that selectively activates the sparse experts, our MoE achieves a 1.28$\times$ speedup in training and 1.23$\times$ speedup in inference. Despite these efficiency gains, we interestingly find that MoE consistently outperforms Multi-E in terms of the cross-dataset detection performance. This suggests that naively aggregating multiple experts may not be the optimal strategy for face forgery detection. One potential explanation for this phenomenon is that using all feature extractor experts could suppress the most informative features while introducing noisy ones. 
The results in Table~\ref{MultiE} further underscores the effectiveness of the proposed method in selecting the optimal expert for face forgery detection.

\begin{figure*}[ht]
\centering
\includegraphics[scale=0.485]{ 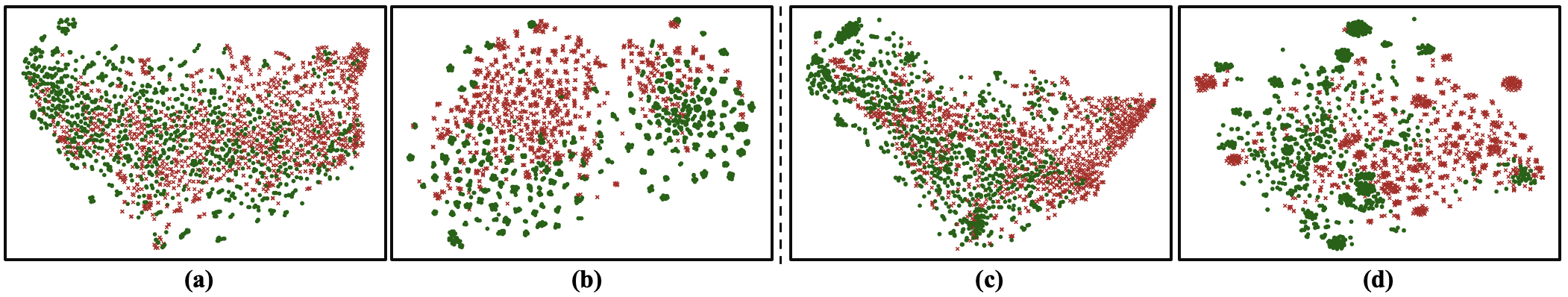}
\caption{t-SNE feature distributions of the ViT baseline and MoE-FFD. Visualization results of (a) ViT baseline on CDF dataset, (b) MoE-FFD on CDF dataset, (c) ViT baseline on WDF dataset, and (d) MoE-FFD on WDF dataset.}
\label{TSNE}
\end{figure*}

\begin{figure*}[ht]
\centering
\includegraphics[scale=0.4]{ 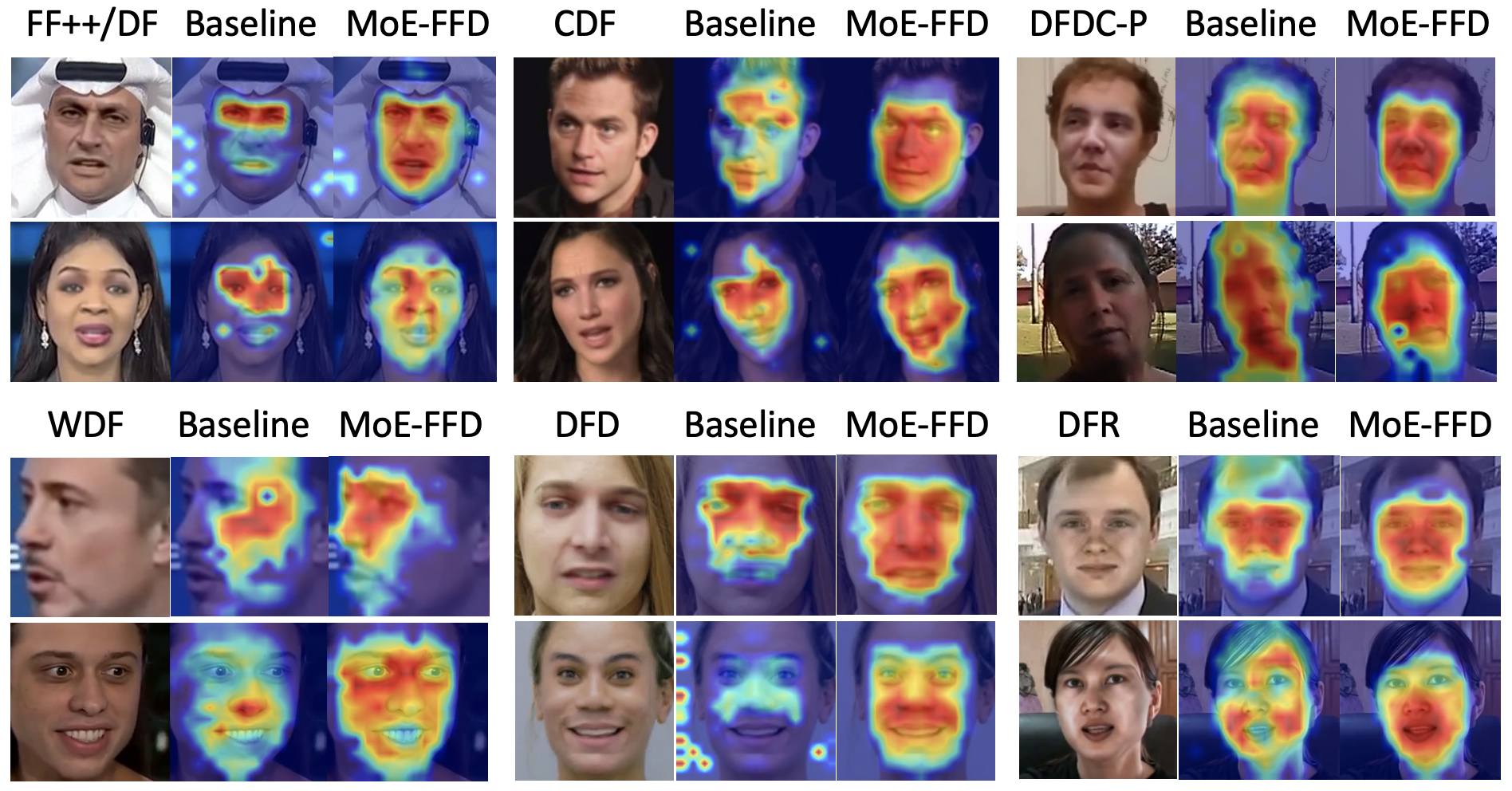}
\caption{Grad-CAM maps of the baseline model (ViT) and our proposed method MoE-FFD on six Deepfake datasets: FF++/DF, CDF, DFDC-P, WDF, DFD, and DFR.}
\label{GRADCAM}
\end{figure*}

\subsubsection{Effectiveness of MoE Loss.} 
The gating network often exhibits a tendency to consistently assign large weights to only a few experts \cite{shazeer2017outrageously}, resulting in overfitting problems. To address this issue, we introduce an MoE loss component aimed at encouraging equal importance among all experts. This regularization also prevents the model from getting trapped in local optima. In this subsection, we examine the impacts of the proposed $L_{moe}$. Table~\ref{loss} presents the cross-dataset evaluation performance with different values of the $L_{moe}$ loss weight $\lambda$ in Eq.~(5), where $\lambda = 0$ represents no MoE loss applied. We observe that the use of $L_{moe}$ effectively mitigates the gate overfitting problem and consistently boosts model generalizability. Furthermore, the model achieves the best generalizability with $\lambda = 1$.

These findings highlight the importance of incorporating balanced expert contributions in the MoE framework. $L_{moe}$ component ensures that the learned representations are more diverse and generalizable across different datasets. This is particularly crucial in practical applications where data distribution can vary significantly. The optimal performance at $\lambda = 1$ suggests that the balance between expert utilization and regularization is vital. 

\subsection{Visualization Results}
To better demonstrate the effectiveness of our MoE-FFD method, we visualize the feature distributions of the baseline model (ViT-B) and MoE-FFD on the CDF and WDF datasets. In Fig.~\ref{TSNE}, green and red marks represent real and fake data samples, respectively. As shown in Fig.~\ref{TSNE} (a) and (c), the baseline model struggles to discriminate the real faces from fake ones, leading to limited detection performance. In contrast, MoE-FFD feature distribution maps in Fig.~\ref{TSNE} (b) and (d) illustrate that the real and fake faces are more discriminative. 

We further provide the Grad-CAM maps of the baseline model and MoE-FFD in Fig.~\ref{GRADCAM}. Both models are trained on FF++ dataset and tested on six Deepfake datasets: FF++/DF, CDF, DFDC-P, WDF, DFD, and DFR datasets. While the baseline model often neglects informative fake facial regions or attends to peripheral irrelevant areas, MoE-FFD consistently directs attention to the manipulated regions within each input face. Despite the diverse environments captured in these datasets, including conditions like poor illumination, extreme head pose, and low resolution, MoE-FFD accurately localizes the forgery regions containing abundant forgery features. This observation further underscores the generalizability of our method.

\begin{table}
  \caption{MoE v.s. Multi-Experts.}
  \label{MultiE}
  \centering
  \renewcommand\arraystretch{1.15}
  \scalebox{0.8}{\begin{tabular}{ccccccccc}
\hline
\multirow{2}{*}{Method} & {Training} & {Inference} & \multicolumn{2}{c}{CDF} & \multicolumn{2}{c}{DFDC-P} & \multicolumn{2}{c}{WDF} \\ \cline{4-9}
& Speed & Speed & AUC & EER & AUC & EER & AUC & EER \\ \hline						
Multi-E & 1.67Iter/s &  7.18Iter/s& 85.04 & 23.74 & 80.16 & 27.78 & 78.67 & 30.07 \\ \hline
MoE & \textbf{2.13}Iter/s & \textbf{8.82}Iter/s & \textbf{86.69} & \textbf{22.06} & \textbf{80.83} & \textbf{26.67} & \textbf{80.64} & \textbf{27.11} \\ \hline
\end{tabular}}
\end{table}

\begin{table}
  \caption{Effectiveness of loss components.}
  \label{loss}
  \centering
  \renewcommand\arraystretch{1.15}
  \scalebox{1.09}{\begin{tabular}{cccccccc}
\hline
\multirow{2}{*}{$L_{ce}$} & \multirow{2}{*}{$\lambda$} & \multicolumn{2}{c}{CDF} & \multicolumn{2}{c}{DFDC-P} & \multicolumn{2}{c}{WDF} \\ \cline{3-8}
& & AUC & EER & AUC & EER & AUC & EER \\ \hline			
\Checkmark &  0 & 82.65 & 25.57 & 78.14 & 29.06 & 76.29 & 30.43\\ \hline
\Checkmark & 0.1 & 85.09 & 23.29 & 80.05 & 26.96&  79.56& 27.92\\ \hline  
\Checkmark & 1 & \textbf{86.69} & \textbf{22.06} & 80.83 & \textbf{26.67} & \textbf{80.64} & \textbf{27.11} \\ \hline
\Checkmark & 5 &  84.01&  24.69&  \textbf{81.47}&  26.68&  80.14&27.32 \\ \hline
\Checkmark & 10 &  83.35&  25.53&  80.46&  28.13&  78.93& 29.05\\ \hline
\end{tabular}}
\end{table}


\section{Conclusions and Future Works}
In this paper, we introduced MoE-FFD, a generalized yet parameter-efficient method for detecting face forgeries. By incorporating external lightweight LoRA and Adapter layers with the frozen ViT backbone, our framework adaptively acquired forgery-specific knowledge with minimal activated parameters. This approach not only harnesses the expressiveness of transformers but also capitalizes on the local forgery priors via customized adapters, thereby significantly enhancing the detection performance. Through dynamic expert selection within both LoRA and Adapter layers, our MoE design further boosts the model's generalizability and robustness. Extensive experiments consistently demonstrated MoE-FFD's superiority in face forgery detection across diverse datasets, manipulation types, and perturbation scenarios. Moreover, MoE-FFD serves as a parameter-efficient detector and can seamlessly adapt to various ViT backbones, facilitating its deployment and fine-tuning in real-world applications. Last but not least, comprehensive ablation studies demonstrated the effectiveness of our designed LoRA layers and Convpass Adapter layers, and the MoE learning scheme indeed helped the model to search the optimal forgery experts. We anticipate that MoE-FFD will inspire future advancements in face forgery detection, particularly in bolstering generalization and efficiency.

While our MoE-FFD framework achieves generalized and robust performance in image-level face forgery detection, extending our network design and learning scheme to video-level Deepfake detection represents an important future research direction. Moreover, detecting audio-visual content forgery is essential in real world applications, multimodal Deepfake detection opens an important research path forward. We envision that our proposed MoE structure and the PEFT strategy should be still effective in these tasks. This path can be explored in our future work.

\ifCLASSOPTIONcaptionsoff
  \newpage
\fi

\bibliographystyle{IEEEtran}
\bibliography{main}

\end{document}